%% file: ZOD_iccv_camready.tex
\documentclass[10pt,twocolumn,letterpaper]{article}

\usepackage{iccv}
\usepackage{times}
\usepackage{epsfig}
\usepackage{graphicx}
\usepackage{amsmath}
\usepackage{amssymb}

\usepackage{booktabs}

\usepackage{url}
\usepackage{multirow}
\usepackage{caption}
\usepackage{subcaption}
\usepackage{pifont}
\usepackage{algorithm}
\usepackage{array}
\usepackage[noend]{algpseudocode}
\usepackage{pifont}
\usepackage[detect-all]{siunitx}
\usepackage{threeparttable}
\usepackage[accsupp]{axessibility}  

\usepackage[breaklinks=true,bookmarks=false]{hyperref}

\usepackage[capitalize]{cleveref}
\crefname{section}{Section}{Sections}
\crefname{section}{Section}{Sections}
\crefname{table}{Table}{Tables}
\crefname{table}{Table}{Tables}
\crefname{figure}{Figure}{Figures}
\Crefname{figure}{Figure}{Figures}

\newcommand{\parsection}[1]{\noindent\textbf{#1}:}

\iccvfinalcopy 

\def\iccvPaperID{8866} 

\ificcvfinal\fi

\begin{document}

\title{Zenseact Open Dataset: A large-scale and diverse multimodal dataset for autonomous driving}

\author{
\vspace{1mm}
Mina Alibeigi$^{\ast}$$^{\dagger}$, William Ljungbergh$^{\ast}$, Adam Tonderski$^{\ast}$,\\
\vspace{1mm}
Georg Hess, Adam Lilja, Carl Lindström, Daria Motorniuk, \\
\vspace{3mm}
Junsheng Fu, Jenny Widahl, Christoffer Petersson\\
Zenseact\\
\tt\small {first.last@zenseact.com}
}

\maketitle
\ificcvfinal\thispagestyle{empty}\fi

\def\thefootnote{$\ast$}\footnotetext{Equal contribution.}
\def\thefootnote{$\dagger$}\footnotetext{Corresponding author.}
\def\thefootnote{\arabic{footnote}}

\input{include_iccv_camready/abstract.tex}

\input{include_iccv_camready/introduction.tex}

\input{include_iccv_camready/related_work.tex}

\input{include_iccv_camready/dataset_description}

\input{include_iccv_camready/dataset_analysis.tex}

\vspace{-1mm}
\input{include_iccv_camready/conclusion.tex}

\vspace{-1mm}
\section{Limitations}
\label{sec:limitations}

While we are excited about ZOD's potential, it's crucial to acknowledge its constraints. The dataset currently offers keyframe annotations, suitable for various tasks (including tasks that require spatiotemporal reasoning) but lacks support for consistent object tracking over time. We plan to enhance this with full sequence annotations. While we've analyzed the effect of anonymization on 2D object detection in ZOD, the impact on other tasks remains unexplored. Our 3D annotations reach up to 250 meters, but their quality and recall wane at these distances. This gap is bridged somewhat by our 2D annotations, which extend much further. Finally, annotations for traffic signs and ego-road are unavailable for some \textit{Frames}. We look forward to improving ZOD with future updates and encourage community feedback.

\section{Acknowledgements}
\label{sec:acknowledgements}
The Zenseact Open Dataset would not be possible without the support of several individuals within our organization. We would like to thank, without any particular order, the following people for their contributions to the dataset: Oleksandr Panasenko, Jakub Bochynski, Dónal Scanlan, Benny Nilsson, Jonas Ekmark, Bolin Shao, Georgios Efthymiou, Erik Rosén, Oleksii Khakhlyuk, Pavel Lutskov, Maryam Fatemi, Joakim Johnander, Mats Nordlund, Joakim Frid, Goran Widborn, and Erik Coelingh.

This work was partially supported by the Wallenberg AI, Autonomous Systems and Software Program (WASP) funded by the Knut and Alice Wallenberg Foundation.

{\small
\bibliographystyle{ieee_fullname}
\bibliography{egbib}
}

\clearpage
\input{include_iccv_camready/new_supplementary.tex}

\end{document}

%% file: include_iccv_camready/abstract.tex
\begin{abstract}
Existing datasets for autonomous driving (AD) often lack diversity and long-range capabilities, focusing instead on 360\si{\degree} perception and temporal reasoning. To address this gap, we introduce Zenseact Open Dataset (ZOD), a large-scale and diverse multimodal dataset collected over two years in various European countries, covering an area 9$\times$ that of existing datasets. ZOD boasts the highest range and resolution sensors among comparable datasets, coupled with detailed keyframe annotations for 2D and 3D objects (up to 245m), road instance/semantic segmentation, traffic sign recognition, and road classification. We believe that this unique combination will facilitate breakthroughs in long-range perception and multi-task learning. The dataset is composed of {\normalfont Frames}, {\normalfont Sequences}, and {\normalfont Drives}, designed to encompass both data diversity and support for spatio-temporal learning, sensor fusion, localization, and mapping. {\normalfont Frames} consist of 100k curated camera images with two seconds of other supporting sensor data, while the 1473 {\normalfont Sequences} and 29 {\normalfont Drives} include the entire sensor suite for 20 seconds and a few minutes, respectively. ZOD is the only large-scale AD dataset released under a permissive license, allowing for both research and commercial use. More information, and an extensive devkit, can be found at \href{https://zod.zenseact.com}{\url{zod.zenseact.com}}.
\end{abstract}
\vspace{-2mm}

%% file: include_iccv_camready/introduction.tex
\section{Introduction}
\label{sec:introduction}

\begin{figure}[t]
    \centering
    \includegraphics[width=0.79\linewidth]{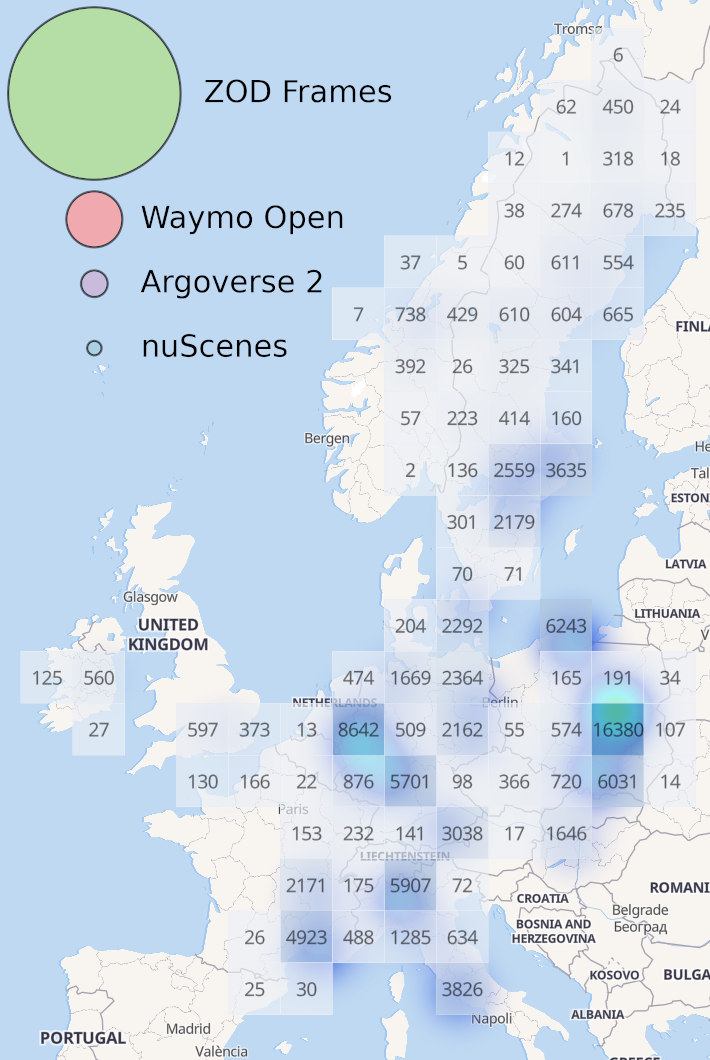}
    \caption{Geographical coverage comparison with other AD datasets using the diversity area metric defined in \cite{waymo_2020} (top left), and geographical distribution of ZOD \textit{Frames} overlaid on the map. The numbers in the quantized regions represent the amount of annotated frames in that geographical region. ZOD \textit{Frames} contain data collected over two years from 14 different European countries, from the north of Sweden down to Italy.}
    \label{fig:geographical-dist}
    \vspace{-5mm}
\end{figure}

Road traffic accidents cause more than 1.3 million deaths and many more nonfatal injuries and disabilities globally each year \cite{road_safety_report}. Automated driving has the potential to improve road safety by intervening in accident-prone situations or even controlling the entire ride from start to destination. Regardless of the level of automation, autonomous vehicles require sensors such as cameras, GNSS (Global Navigation Satellite System), IMU (Inertial Measurement Unit), and range sensors such as radar or LiDAR (Light Detection and Ranging) to perceive their surroundings accurately. Moreover, they require advanced perception, fusion, and planning algorithms to make use of this data efficiently. Machine learning (ML) algorithms, particularly deep learning (DL), have been increasingly used to develop autonomous driving (AD) software, but they require high-quality and diverse data from real-world traffic scenarios to achieve the necessary performance.

The development of AD systems owes much of its recent success to the availability of large-scale image datasets~\cite{camvid_2008, cityscape_cvpr16, imagenet_cvpr09, coco_eccv14, vistas_iccv17, bdd100k_2018} and dedicated multimodal AD datasets~\cite{nuscenes_cvpr_2020, argoverse1_2019, kitti_2013_IJRR, a2d2_2020, once_2021, waymo_2020, argoverse2_2021, xiao2021pandaset}. These AD datasets have emphasized temporal reasoning and 360\si{\degree} perception, as this corresponds to a nominal self-driving setup. However, as data from the same scene is highly correlated, this naturally limits diversity in terms of weather or lighting conditions, driving situations, and geographical distribution. This can result in overly specialized solutions, which may not generalize to the full operational design domain of real-world AD systems.

To complement existing datasets, we introduce Zenseact Open Dataset (ZOD), Europe's largest and most diverse multimodal AD dataset. ZOD consists of more than 100k traffic scenes that have been carefully curated to cover a wide range of real-world driving scenarios. The dataset is split into three subsets: 100k independent \textit{Frames}, 1473 twenty-second \textit{Sequences}, and 29 \textit{Drives} lasting a few minutes. \textit{Frames} are primarily suitable for non-temporal perception tasks, \textit{Sequences} are intended for spatio-temporal learning and prediction, and \textit{Drives} are aimed at longer-term tasks such as localization, mapping, and planning. This separation allows ZOD to cover an area 9$\times$ larger than any other AD dataset, offering ample opportunities for developing robust algorithms that generalize well across multiple operational domains. We also facilitate research in domain adaptation and transfer learning by providing comprehensive metadata for each scene.

Robust performance in various conditions, including high speeds, will be vital for the deployment of AD systems. In particular, high-speed scenarios puts a hard requirement on long-range perception, which in turn puts challenging requirements on sensor resolution. ZOD stands out from other AD datasets by employing high-resolution sensors, such as an 8MP camera, rooftop LiDARs with 254k points per scan, and a high-precision GNSS/IMU inertial navigation system with 0.01\si{\meter} position accuracy. Additionally, we provide manual keyframe annotations for several perception tasks, such as semantic and instance segmentation masks for roads and lanes, 2D and 3D bounding boxes for static and dynamic objects (up to 245 meters), and road condition labels. We further annotate traffic signs with a rich taxonomy of 156 classes. The 446k unique labeled instances constitute the largest traffic sign dataset to date. We believe that the combination of high-quality sensors and detailed annotations in ZOD will enable breakthroughs in accurate and long-range perception, which is crucial for high-speed driving scenarios.

ZOD's extensive annotations for multiple perception tasks make it an ideal dataset for multi-task learning, which is a recent trend in computer vision and AD \cite{multitask_dense_2022,multitask_overview_2017, multitask_survey_2022}. The core idea of multi-task learning is to learn a shared representation that can benefit all tasks, resulting in improved generalization and performance on individual tasks \cite{multitasklearning_caruana}. This approach also allows models to make better use of data and available resources, a crucial feature for real-world applications such as AD systems as they typically operate on embedded hardware with limited computational power. 

To ensure the privacy of individuals and comply with legal and regulatory requirements, we employ two approaches to anonymize faces and license plates: blurring and replacement with synthetic data. These anonymization techniques were chosen to enable research on the impact of anonymization techniques on learning methods, with initial results demonstrating that none of the techniques have a negative impact on performance. 

Finally, ZOD is the first large-scale AD dataset released under the permissive CC BY-SA 4.0 license \cite{License}. This license allows for research and commercial use (subject to the license terms), as well as sharing and adapting permits, which provides an opportunity for startups and other commercial entities to leverage the dataset for their projects. We believe that this open and inclusive approach will foster innovation and accelerate the development of AD technology beyond the research community. To facilitate a rapid start with ZOD, it comes with an extensive development kit, including multiple tutorials and examples.

In summary, our main contributions are the following:

\begin{itemize}
    \item We release ZOD, the most diverse autonomous driving dataset to date. The data is collected from Europe over multiple years and is curated to contain a wide range of traffic scenarios, weather conditions, road types, and lighting conditions.
    \item The data is collected using high-resolution sensors and coupled with detailed keyframe annotations for 2D/3D objects, lane instances, and road segmentation, enabling long-range perception with annotated objects farther away than in any other comparable AD dataset.
    \item The object annotations include a rich traffic sign taxonomy with more than twice as many unique instances as the largest existing traffic sign dataset. 
    \item ZOD is the first large-scale AD dataset released under a permissive license, allowing both research and commercial use.
\end{itemize}

%% file: include_iccv_camready/related_work.tex
\newcommand{\sqkm}{\si{\kilo\metre\squared}}
\begin{table*}[t]
    \centering
    \resizebox{\textwidth}{!}{%
        \begin{tabular}{l|c|c|c|c|c|c|c|c|c}
            Dataset                             & Locations                      & Geo. coverage                  & Ann. frames             & Sequences     & Size (hr)                     & $\text{Ann. range}^{\dagger}$        & Avg. LiDAR points & Camera        & Map          \\ \hline
            KITTI \cite{kitti_2013_IJRR}        & Karlsruhe                      & -                              & 15k$^{\star}$           & 22            & 1.5                           & 91 \si{\metre}                       & 120k              & $90^{\circ}$  & No           \\
            nuScenes \cite{nuscenes_cvpr_2020}  & Boston, Singapore              & 5 \sqkm$^{\mathsection}$       & 40k$^{\star}$           & 1000          & 5.5                           & 141 \si{\metre}                      & 34k               & $360^{\circ}$ & \textbf{Yes} \\
            ONCE    \cite{once_2021}            & China                          & -                              & 16k$^{\star}$           & 581           & \textbf{27.8}                 & 81 \si{\metre}                       & 65k               & $360^{\circ}$ & No           \\
            PandaSet    \cite{xiao2021pandaset} & San Francisco                  & -                              & 8k$^{\star}$            & 103           & 0.2                           & \textbf{300} {\bfseries \si{\metre}} & 166k              & $360^{\circ}$ & No           \\
            Waymo Open \cite{waymo_2020}        & 6 U.S. cities                  & 76  \sqkm$^{\mathsection}$     & \textbf{400k}$^{\star}$ & \textbf{2030} & 11.3                          & 80 \si{\metre}                       & 177k              & $360^{\circ}$ & No           \\
            A2D2  \cite{a2d2_2020}              & 3 German cities                & -                              & 12k                     & -             & -                             & 103 \si{\metre}                      & 7k                & $360^{\circ}$ & No           \\
            Argoverse 2  \cite{argoverse2_2021} & 6 U.S cities                   & 17  \sqkm                      & 150k$^{\star}$          & 1000          & 4.2                           & 214 \si{\metre}                      & 107k              & $360^{\circ}$ & \textbf{Yes} \\
            \midrule
            ZOD \textit{Frames}                 & \textbf{14 European countries} & \textbf{705} {\bfseries \sqkm} & 100k                    & -             & $\textbf{55.6}^{\ddagger}$  & 245  \si{\metre}                     & \textbf{254k}     & $120^{\circ}$ & No           \\
            ZOD \textit{Sequences}              & 6 European countries           & 26 \sqkm                       & 1473                    & 1473          & 8.2                           & 245  \si{\metre}                     & \textbf{254k}     & $120^{\circ}$ & No           \\
            ZOD \textit{Drives}                 & 2 European countries           & -                              & -                       & 29            & 1.5                           & -                                    & \textbf{254k}     & $120^{\circ}$ & No           \\
        \end{tabular}
    }
    \caption{Dataset comparison. Geographical coverage is computed over annotated frames, $^{\mathsection}$ refers to values taken from \cite{waymo_2020}. $^{\star}$Sequential annotation, allowing temporal tasks. $^{\ddagger}$Counting each \textit{Frame} as a 2-second LiDAR sequence, note that only the center image is provided. $^{\dagger}$Showing the 99.9th percentile computed using the publicly available data. }
    \label{tab:dataset-comparison}
\end{table*}

\section{Related work}
\label{sec:related_work}

Over the last decade, the development of AD datasets has been a focus area to advance AD research and enhance road safety. Many of these datasets have been devoted to visual perception \cite{camvid_2008, cityscape_cvpr16, vistas_iccv17, bdd100k_2018}; However, in recent times, multimodal datasets have become increasingly popular as most AD systems aim to fuse information from various onboard sensors to generate a robust representation of the world and make more informed decisions. 

KITTI \cite{kitti_2013_IJRR} is widely regarded as one of the most impactful multimodal AD datasets to date. Released in 2012, KITTI provides 22 road sequences with stereo cameras, LiDAR, and high-precision GNSS/IMU sensor data from real-world driving in Karlsruhe, Germany. With 200k object labels in the form of 3D tracklets, KITTI enabled significant advancements in AD research, including 3D object detection and tracking, visual odometry, and scene flow estimation. However, as modern algorithms and AD systems tackle more complex tasks, there is a growing demand for larger, more diverse, and more advanced datasets.

Since the release of KITTI, the field of AD research has seen several large-scale AD datasets striving to push the boundaries in various aspects. Among these is the ApolloScape dataset \cite{apolloscape_2019}, which boasts one of the largest publicly available collections of annotated video frames for semantic segmentation in AD, with over 140k frames. However, the dataset's strong temporal connection between frames limits the diversity of training data. The KAIST multispectral dataset \cite{KAIST_2018} draws attention for its use of thermal cameras, while H3D \cite{H3D_2019} stands out as one of the first datasets to fully annotate 3D objects in a \si{360\degree} perspective. A2D2 \cite{a2d2_2020} is one of the first public AD datasets allowing commercial use but with restrictions on sharing the derivates. It offers RGB images, LiDAR data, and vehicle bus data, with 41k frames containing semantic segmentation labels and 12k frames with 3D bounding box labels for objects visible in the front camera's field of view. However, A2D2 suffers from extremely sparse point clouds. PandaSet~\cite{xiao2021pandaset} is also released under a permissive license, and thanks to their multi-LiDAR setup, it offers much denser point clouds which allow them to annotate 3D bounding boxes up to 300m. Similar to A2D2, they also offer RGB images and vehicle bus data, together with annotations across their 8k frames. While being multi-modal datasets, both A2D2 and PandaSet suffer from low annotation counts, and similarly to H3D and KAIST, their limited size has hindered widespread adoption.

NuScenes \cite{nuscenes_cvpr_2020}, released in 2019, became a significant milestone in AD research as one of the first publicly available large-scale datasets. It provides a rich surround view by means of comprehensive sensor data from LiDAR, six cameras, and five radars. Additionally, the dataset includes semantic maps of roads and sidewalks, vehicle bus data, and GNSS information. NuScenes is organized into sequences of 20 seconds and provides detailed temporal annotations for 3D objects at \si{2\hertz}, making it one of the first datasets to do so. This feature has enabled many breakthroughs in the development of algorithms for detection, tracking, prediction, and planning.

Waymo Open Dataset~\cite{waymo_2020} and Argoverse 2~\cite{argoverse2_2021} are two large-scale datasets that adopt the strengths of nuScenes, while addressing many of its weaknesses. Both datasets cover a more comprehensive range of driving scenarios and include higher-resolution sensors. Waymo's dataset excels in geographical diversity, while Argoverse 2 stands out for its long-range annotations, maps, and extensive object taxonomy. While these three datasets push the state-of-the-art in AD, especially in terms of spatio-temporal reasoning and surround vision, they do so at the cost of high sample correlation.

\begin{table*}[t]
  \centering
  \begin{threeparttable}
  \begin{tabular}{p{0.12\linewidth} | p{0.88\linewidth}} 
    \toprule
    Sensors & Details \\
    \midrule
    LiDARs &
    1xVelodyne VLS128 rotating 3D laser scanner, HFOV\tnote{*} ~360\si{\degree}, VFOV\tnote{*}~ 40\si{\degree} [-25\si{\degree}, +15\si{\degree}], HRES\tnote{*}~ 0.1\si{\degree} to 0.4\si{\degree}, VRES\tnote{*}~ 0.11\si{\degree}, channels 128, wavelength \si{903 \nano\meter}, range up to \si{245 \meter}, and frame rate \si{10 \hertz}. \\
    & 2xVelodyne VLP16, HFOV 360\si{\degree}, VFOV 30\si{\degree} [-15\si{\degree}, +15\si{\degree}], HRES 0.1\si{\degree} to 0.4\si{\degree}, VRES 2\si{\degree}, channels 16, wavelength \si{905 \nano\meter}, range up to \si{100 \meter}, frame rate \si{10 \hertz}, collecting up to 0.3 million points/second.
    \\
    \midrule
    Camera & 1xRGB front-looking camera, HFOV 120\si{\degree}, VFOV 67\si{\degree}, and resolution 3848x2168 (8MP).
    \\
    \midrule
    High precision GNSS/IMU & 1xOxTS RT3000 inertial and GNSS navigation system, six axes, L1/L2 RTK with frame rate \si{100 \hertz}, 0.01\si{\meter} position accuracy, 0.03\si{\degree} pitch/roll and 0.1\si{\degree} heading accuracy.\\
    \bottomrule
  \end{tabular}
  \begin{tablenotes}\footnotesize
    \item[*] HFOV/VFOV and HRES/VRES represent the horizontal/vertical field of view and resolutions, respectively.
  \end{tablenotes}
  \caption{Sensor specifications of ZOD.}
  \label{tab:sensor-setup}
  \end{threeparttable}
\end{table*}

In contrast, ZOD makes a different set of trade-offs. Our \textit{Frames} subset has nine times larger geographical coverage than the Waymo Open Dataset, with only a quarter of the frames. By complementing existing datasets, our goal is to facilitate breakthroughs in long-range perception and multi-task learning and enable the benchmarking of algorithms for both research and commercial use.

Another recent trend in the AD community is to release large-scale datasets without annotations, primarily for self-supervised learning purposes. The ONCE dataset~\cite{once_2021} is one such example, containing a million LiDAR scenes and their corresponding camera images collected over three months in various areas, lighting, and weather conditions throughout China. However, the LiDAR point clouds in ONCE are limited to 65k points per frame, and the dataset lacks localization information, such as GNSS or map data. Similarly, the Argoverse 2 Lidar~\cite{argoverse2_2021} dataset contains a vast number of unlabeled LiDAR sequences; however, it does not include camera data, which is essential for many AD tasks. To this end, ZOD includes \textit{Sequences} and \textit{Drives} subsets that are mostly unlabeled and consist of diverse traffic scenes with the camera, LiDAR, high-precision, and consumer-grade GNSS/IMU data.

Traffic sign recognition is another essential research avenue for enabling AD systems, that is typically pursued separately. The German Traffic Sign Benchmark Dataset \cite{stallkamp2011gtrsb} is one of the first datasets created to facilitate research on the classification of traffic signs. Multiple regional datasets \cite{larsson2011sts, mathias2013btsd, shakhuro2016rtsid, zhu2016tt100k} have followed since, containing a varying number of images, signs, and classes. The Mapillary Traffic Sign Dataset (MTSD) \cite{ertler2020mapillary} is one of the largest and most diverse traffic sign datasets to date, containing 206k labeled sign instances and a taxonomy of 313 classes from all over the world. Although ZOD has a smaller taxonomy of 156 classes, it has twice the amount of labeled sign instances (\ie, 446k) compared to MTSD.

We refer the reader to \cref{tab:dataset-comparison} for a comprehensive comparison of ZOD with other multimodal AD datasets.

%% file: include_iccv_camready/dataset_description.tex
\section{Zenseact Open Dataset}
\label{sec:dataset_description}

ZOD is a multimodal dataset containing a variety of real-world traffic scenes from highways, urban areas, and country roads around Europe. The dataset is collected under diverse weather conditions (clear, cloudy, rainy, and snowy) and lighting conditions (day, night, and twilight) over two years. It contains an extensive and detailed collection of fine-grained annotations for various tasks, including semantic and instance segmentation for the road, 2D and 3D bounding boxes for the dynamic and static objects including traffic signs with a rich taxonomy, and road classification labels.

The subsequent sections provide an elaborate description of the sensor suite, data, and annotations, followed by a statistical analysis of the dataset in \cref{sec:dataset_analysis}.

\begin{figure}[b]
  \centering
   \includegraphics[width=\linewidth]{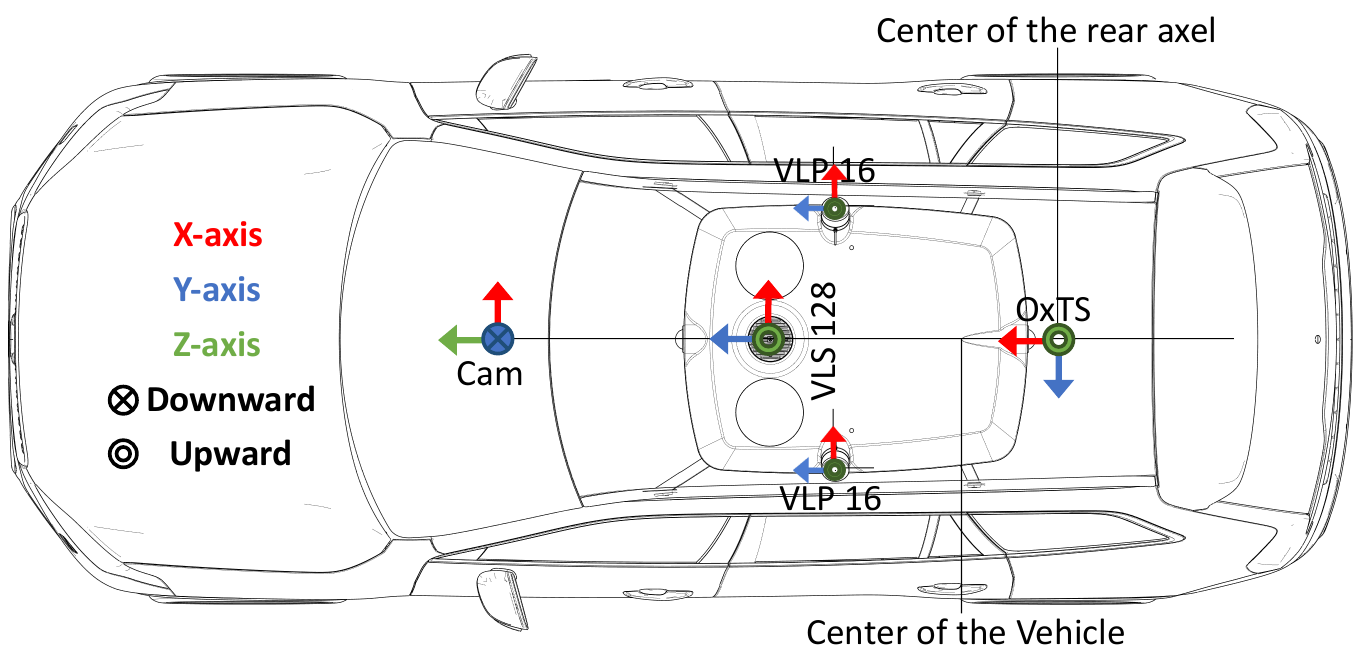}
   \caption{Placement of sensors used by data collection vehicles in ZOD and their corresponding coordinate systems.}
   \label{fig:sensor-setup}
\end{figure}

\begin{figure*}[tb]
    \centering
    \includegraphics[width=\linewidth, trim={1mm, 0, 1mm, 0}, clip]{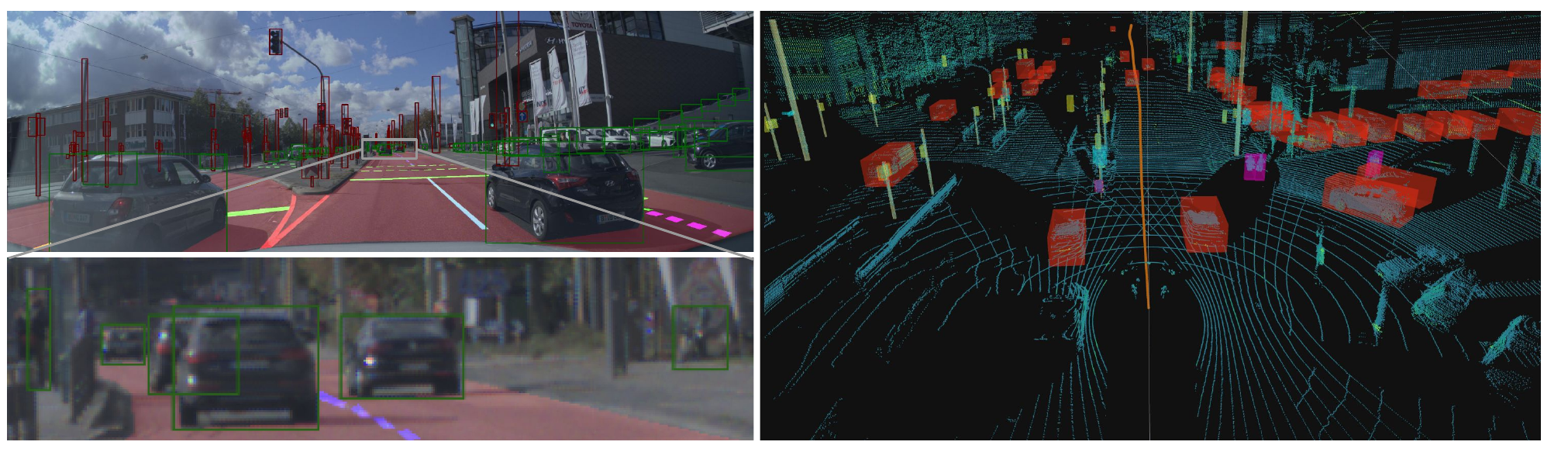}
    \label{fig:oxts_lidar_projection}
     \caption{One \textit{Frame} overlaid with multi-task annotations (top-left) and a zoomed-in version (bottom-left), highlighting annotated dynamic objects and lane instances at long distances. We also show the LiDAR point cloud with annotated 3D boxes and 200 \si{\meter} of future trajectory for the ego-vehicle (right) from the same \textit{Frame}.}
    \label{fig:overlaid-annotations}
\end{figure*}

\subsection{Sensor suite}
\label{sec:sensor-suite}
The data collection has been conducted using several vehicles with an identical sensor layout driven around Europe over the course of two years. The cars are equipped with a high-resolution camera, front- and side-looking LiDARs, and high-precision GNSS/IMU sensors. \cref{fig:sensor-setup} illustrates the sensor locations and their respective coordinate systems, whereas \cref{tab:sensor-setup} gives comprehensive sensor specifications.

\parsection{LiDAR data}
The LiDAR point clouds are captured at \si{10 \hertz} and stored in a standard binary file format (\texttt{.npy}) per scan. Each file contains data from all three LiDAR sensors (VLS128 and VLP16s), represented as 
a 6-dimensional vector with the timestamp, 3D coordinates (x, y, and z), intensity, and diode index. The timestamp is relative to the frame timestamp in UTC, and the 3D coordinates are in meters. Intensity is a measure of the reflection magnitude ranging from 0 to 255, and the diode index specifies the emitter that produced the point. Each LiDAR point cloud contains around 254k points on average and can be easily read and visualized using the provided development kit.

\parsection{High-precision GNSS/IMU data}
The high-precision GNSS/IMU data is logged at \si{100\hertz} and stored as HDF5 files, including UTC timestamp in seconds, WGS84 geographic coordinates (latitude, longitude, and altitude), ECEF Cartesian coordinates, heading, pitch, roll, velocities, accelerations, angular rates, and satellite information. This data can be used as ground truth for training different ML models, such as ego-vehicle trajectory prediction. A comprehensive description of the fields, coordinate transformation, and visualization functionalities
(like the ego-vehicle trajectory in \cref{fig:overlaid-annotations}) can be found in the development kit.

\parsection{Camera data}
\label{sec:camera-data}
The camera data is captured by high-resolution (8MP) wide-angle fish-eye lenses. All raw captured camera data is converted to RGB images using an internal production-level image signal processor. The RGB camera images are captured at \si{10\hertz} and saved as JPG files for a more accessible download of ZOD. However, we also provide lossless-PNG camera images. Considering the marginal compression impact on the learning models (see the supplementary material for empirical evidence), we strongly suggest using JPG images for benchmark experiments on ZOD.

\parsection{Vehicle data}
Various vehicle data are also released for \textit{Sequences} and \textit{Drives}. These include vehicle control signals such as steering wheel angle, acceleration/brake pedal ratios, and turn indicator status, as well as consumer-grade IMU and satellite positioning data. The vehicle control signals, IMU, and satellite positioning data are logged at \si{100\hertz}, \si{50\hertz}, and \si{1\hertz}, respectively.

\subsection{Calibration and coordinate systems}
To avoid drift over time and to achieve good cross-modality data alignment, all sensors are carefully and regularly synchronized and calibrated with regard to the specified ISO-8855 reference coordinate system during the data collection process. The origin of the reference coordinate system is at a fixed point relative to the vehicle chassis such that it appears in the center of the rear axle, given identical load conditions as during calibration. Under these conditions, it has the axes $X$-forward, $Y$-left, and $Z$-up. Individual sensor calibrations are provided for each datapoint, containing all the required information (\eg, intrinsic and extrinsic calibration) to transform data between any two sensors. Moreover, functionalities for coordinate transformations and projections are available in the development kit.
\subsection{Privacy protection}
To protect the privacy of every individual in our dataset, and to comply with privacy regulations such as the European Union’s General Data Protection Regulation (GDPR) \cite{voigt2017eu}, we use third-party services \cite{BrighterAI} and anonymize all objects in the images that contain personally identifiable information, \ie, human faces and vehicle license plates.

Two anonymized RGB images are provided per \textit{Frame} in ZOD, one using blurring and one using Deep Neural Anonymization (DNAT), see \cref{fig:overlaid-annotations} for examples of license plate anonymization using DNAT. The latter is based on generative AI, has minimal pixel impact, and maintains information like the line of sight of pedestrians while preserving anonymity. To the best of the authors’ knowledge, ZOD is the only dataset that provides two anonymized versions of the original camera images, enabling an impact study of anonymization approaches on the quality of ML models (\cref{sec:anonymization_study}) while supporting use cases such as human intent prediction in a compliant way.

\subsection{Data categories}
ZOD is categorized into three groups: \textit{Frames}, \textit{Sequences}, and \textit{Drives}. The following subsections describe the content of each category, and \cref{sec:sensor-suite} gives details per sensor data.

\parsection{Frames}
We carefully curate and select 100k frames from all over Europe, representing diverse traffic, location, weather, and lighting conditions.
Each \textit{Frame} scene contains two anonymized versions of an RGB camera image (\ie, blurred and DNAT), captured from a front-looking camera mounted at the top of the windshield. We define the camera images as \textit{keyframes}, considering they are fully annotated for different tasks (\cref{sec:data_annotations}). Moreover, a \(\pm\)1-second sequence of LiDAR scans at \si{10\hertz} frequency is provided around each \textit{keyframe}. This has been complemented by adding high-precision GNSS/IMU data at a frequency of \si{100\hertz}, which covers five seconds before and either 25 seconds after the \textit{keyframe} or \si{300\meter} ahead, whichever occurs first. To facilitate the extraction of interesting custom scenarios, a list of metadata describing the timestamp, geographical position, country code, weather conditions (\eg, clear, rainy, foggy, snowy), solar elevation angle, road type (\eg, highway, city), and the total number of annotated objects (\eg, vehicles, pedestrians, vulnerable vehicles) is also provided for each \textit{Frame}.

\parsection{Sequences}
We select an additional 1473 varied scenes, named \textit{Sequences}, each with a duration of 20 seconds. \textit{Sequences} can be utilized in applications that require temporal reasoning, such as visual odometry and ego-vehicle trajectory prediction. After the anonymization impact study on \textit{Frames} (\cref{sec:anonymization_study}), we release only the blurred anonymized version of RGB camera images recorded at \si{10\hertz} for each sequence.
Similar to \textit{Frames}, LiDAR scans and high-precision GNSS/IMU data are also provided per scene, but for the entire 20 seconds. We also offer vehicle data for each \textit{Sequence} scene in ZOD. This could be of particular interest to robotics research and higher-level scene understanding applications.
Furthermore, the middle frame in each \textit{Sequence} scene is carefully annotated for different tasks (\cref{sec:data_annotations}), enabling spatio-temporal learning and automatic annotation generation tasks, among others.

\parsection{Drives}
We provide 29 \textit{Drive} scenes, spanning a few minutes each, from two different cities and contain the same sensor data as the \textit{Sequence} category. \textit{Drive} scenes are deliberately chosen to capture changing road structures (\eg, straight, curvy, banked, T-crossings, roundabouts, splits, merges, on-ramps, off-ramps), different road types (\eg, urban, suburban, highways), and various traffic scenarios (\eg, lane changes, cut-ins) to represent real-world complexities. They also encompass a couple of loop closures. The \textit{Drives} aim to be useful for research areas such as visual-based localization or simultaneous localization and mapping.

\begin{figure}
    \centering
    \includegraphics[width=\linewidth]{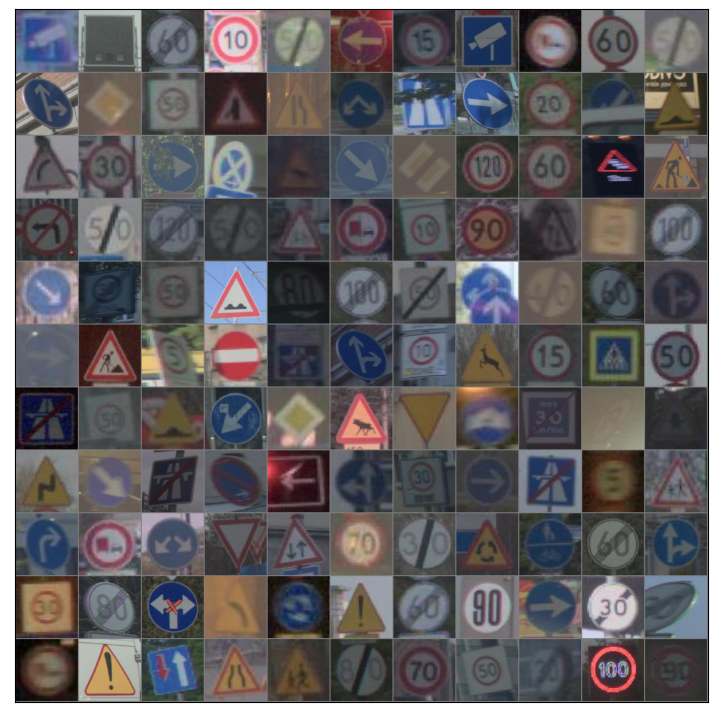}
    \caption{A random selection of cropped traffic signs, extracted from \textit{Frames}. ZOD contains many difficult cases caused by occlusion, distance, viewing angle, lighting, \etc. }
    \label{fig:tsr}
\end{figure}

\begin{figure*}[ht]
\centering
\begin{subfigure}{0.33\linewidth}
    \centering
    \includegraphics[width=\linewidth]{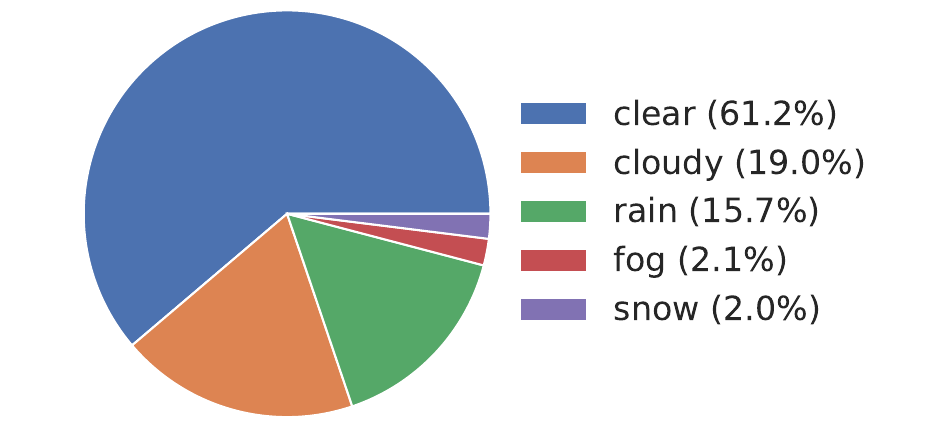}
    \caption{Weather}
    \label{fig:metadata_weather}
\end{subfigure}
\begin{subfigure}{0.33\linewidth}
    \centering
    \includegraphics[width=\linewidth]{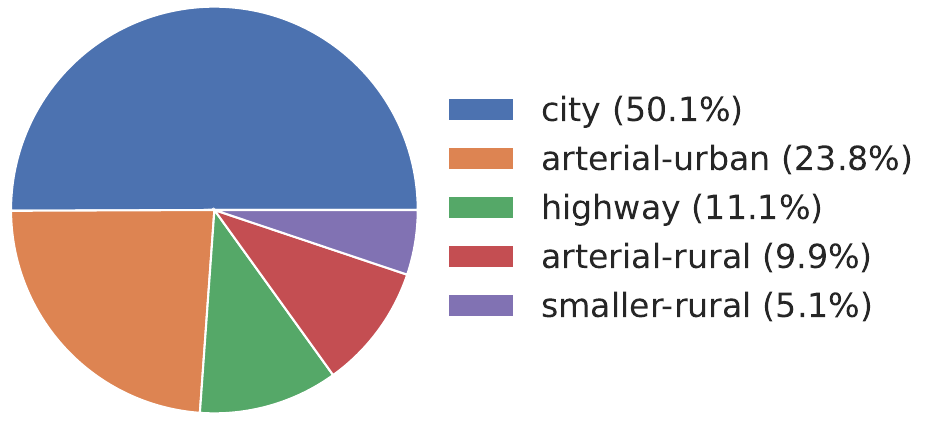}
    \caption{Road type}
    \label{fig:metadata_roadtypes}
\end{subfigure}
\begin{subfigure}{0.33\linewidth}
    \centering
    \includegraphics[width=\linewidth]{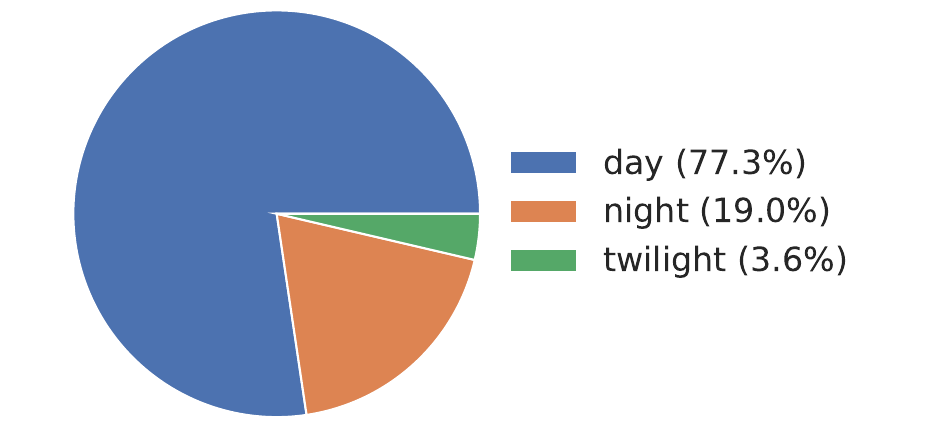}
    \caption{Time of day}
    \label{fig:metadata_daytimes}
\end{subfigure}

    \caption{Distribution of weather (a), road types (b), and time of day (c) in ZOD \textit{Frames}. }
    \label{fig:metadata_weather_roadtype_daytime}
\end{figure*}

\begin{figure*}[ht]
    \centering
    \includegraphics[width=\textwidth]{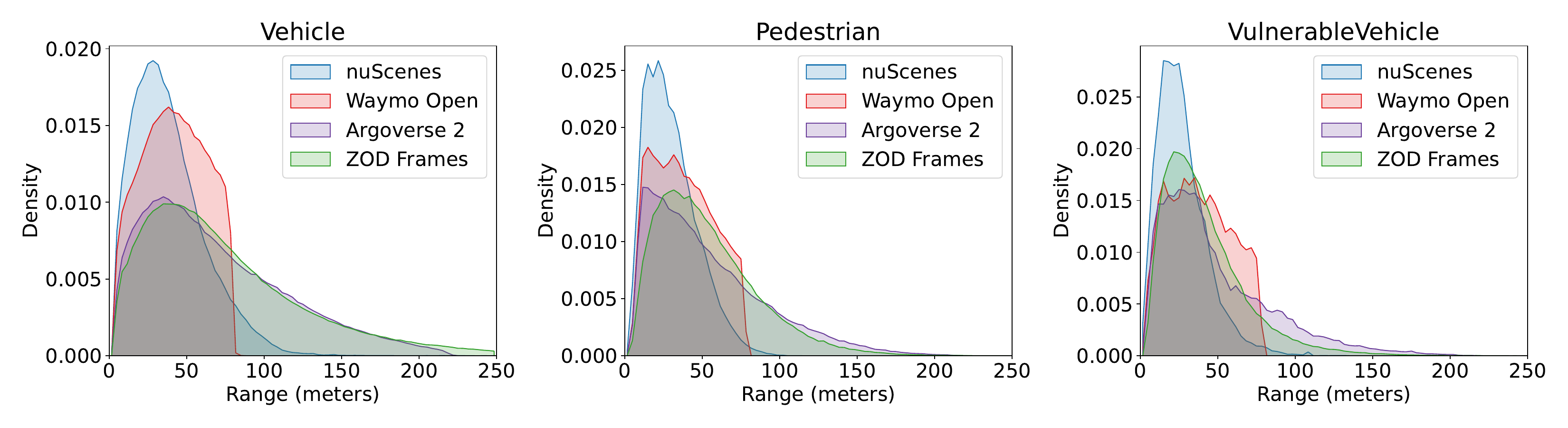}
    \caption{Distribution of the distance to the ego-vehicle over all annotated objects in nuScenes, Waymo Open, Argoverse 2, and ZOD \textit{Frames}. The distribution of ZOD \textit{Frames} is on-par with Argoverse 2. However, the annotated objects in ZOD \textit{Frames} are less temporally correlated.}
    \label{fig:object-range-distribution}
\end{figure*}

\subsection{Annotations}
\label{sec:data_annotations}
We provide large-scale, high-quality, and detailed ground truth labels for several AD tasks per each \textit{keyframe} in the \textit{Frame} and \textit{Sequence} scenes. All labels are manually created by skilled human annotators using commercial labeling tools and further passed through quality checks.
The annotations are separated into three main categories: 1) semantic/instance segmentation masks for lane markings, road paintings, and ego road (a.k.a. driveable area), 2) 2D/3D bounding box labels for dynamic and static objects including traffic signs, and 3) classification labels for the road surface. 

The first category of annotations consists of pixel-wise semantic segmentation labels with 15 top-level classes, see the supplementary material for the exact taxonomy. Instance segmentation labels are also provided for the annotated lane markings, along with additional properties such as color and cardinality, \ie, if a lane marking is single or part of a group of lane markings. \cref{fig:overlaid-annotations} illustrate examples of lane marking and ego road annotations overlaid on the camera image.

All static and dynamic objects present in the camera images are annotated with a tightly fitting 2D bounding box indicated by the pixel coordinates of its four outermost points.
Objects visible both in the camera image and the LiDAR point cloud are also labeled with a 9-DOF 3D bounding box, described by the coordinate of the center of the box, length, width, height size, and the four quaternion rotation parameters of the cuboid (\ie, qw, qx, qy, and qz). ZOD uses a hierarchical taxonomy, where dynamic objects are classified into four higher-level classes, which are further broken down into 16 subclasses. Static objects are classified into seven higher-level classes, which are broken down into 13 subclasses. Traffic signs are treated specially and are further labeled using an extensive taxonomy with 6 larger categories and 156 granular classes (\cref{fig:tsr}). By using this hierarchical taxonomy, ZOD provides a detailed annotation schema that enables researchers to analyze the dataset at different levels of granularity. Furthermore, objects are assigned properties, some generic (\eg, occlusion rate), and some class-specific (\eg, is\_electronic for traffic signs or emergency for vehicles).
\cref{fig:overlaid-annotations} shows examples of the annotated dynamic and static objects.

Lastly, we also provide details of the ego road surface condition (\ie, wet or covered with snow) as road classification labels. Exact annotation taxonomies are detailed in the supplementary material.

%


%% file: include_iccv_camready/dataset_analysis.tex
\section{Dataset analysis} 
\label{sec:dataset_analysis}
In the following subsections, we analyze ZOD and highlight some key characteristics, namely its diversity (\cref{sec:diversity}) and long-range objects (\cref{sec:long-range-objects}), and compare these with existing datasets. Moreover, we show the long-tailed nature of our dataset (\cref{sec:long-tail-perception}) and, lastly, we analyze the impact that different anonymization techniques have on downstream computer vision tasks (\cref{sec:anonymization_study}).

\begin{table*}[ht]
\centering
\resizebox{\textwidth}{!}{%
\begin{tabular}{lcccccccccc}
    \toprule
    Pipeline                        & Mode     & $AP$       & $AP50$ & $AP75$ & $AP_s$ & $AP_m$ & $AP_l$ & $AP_{veh}$ & $AP_{VV}$ & $AP_{ped}$ \\
    \midrule
    \multirow{3}{30pt}{Faster-RCNN}         
        & original & 30.23 $\pm$ 0.09 & 54.79 $\pm$ 0.06 & 28.72 $\pm$ 0.15 & 7.23 $\pm$ 0.04 & 30.49 $\pm$ 0.14 & 51.23 $\pm$ 0.07 & 42.41 $\pm$ 0.07 & 25.96 $\pm$ 0.15 & 22.32 $\pm$ 0.04 \\ 
        & DNAT & 30.28 $\pm$ 0.03 & 54.86 $\pm$ 0.09 & 28.82 $\pm$ 0.08 & 7.15 $\pm$ 0.13 & 30.57 $\pm$ 0.06 & 51.31 $\pm$ 0.08 & 42.48 $\pm$ 0.02 & 25.90 $\pm$ 0.14 & 22.44 $\pm$ 0.03 \\ 
        & blur & 30.17 $\pm$ 0.06 & 54.66 $\pm$ 0.10 & 28.83 $\pm$ 0.08 & 7.24 $\pm$ 0.02 & 30.50 $\pm$ 0.10 & 51.08 $\pm$ 0.10 & 42.41 $\pm$ 0.04 & 25.87 $\pm$ 0.13 & 22.22 $\pm$ 0.04 \\ 
    \midrule
    \multirow{3}{30pt}{YOLOv7}   & original & 33.62 $\pm$ 0.04 & 62.85 $\pm$ 0.03 & 30.84 $\pm$ 0.11 & 13.39 $\pm$ 0.09 & 35.81 $\pm$ 0.03 & 47.02 $\pm$ 0.21 & 47.79 $\pm$ 0.02 & 25.51 $\pm$ 0.04 & 27.56 $\pm$ 0.10 \\
        & DNAT & 33.74 $\pm$ 0.09 & 62.95 $\pm$ 0.03 & 31.10 $\pm$ 0.11 & 13.54 $\pm$ 0.08 & 35.92 $\pm$ 0.14 & 47.17 $\pm$ 0.17 & 47.85 $\pm$ 0.11 & 25.67 $\pm$ 0.11 & 27.71 $\pm$ 0.08 \\
        & blur & 33.67 $\pm$ 0.01 & 62.91 $\pm$ 0.07 & 30.90 $\pm$ 0.01 & 13.51 $\pm$ 0.06 & 35.92 $\pm$ 0.03 & 47.03 $\pm$ 0.02 & 47.89 $\pm$ 0.04 & 25.58 $\pm$ 0.12 & 27.53 $\pm$ 0.05 \\
    \bottomrule
\end{tabular}
}
\caption{Impact of image anonymization. We report AP (computed according to the COCO evaluation protocol \cite{coco_eccv14}) when training image-based object detectors on images anonymized using three separate methods: None (original), DNAT, and blurring, while evaluation is done using the original images. The results show the mean and standard deviation across three separate runs. $AP_{veh}$, $AP_{VV}$, and $AP_{ped}$ refer to $AP$ for the vehicle, vulnerable vehicle, and pedestrian classes, respectively.}
\label{tab:2d-anonymization}
\end{table*}

\subsection{Diversity}
\label{sec:diversity}
Most AD datasets are collected from a handful of cities, see \cref{tab:dataset-comparison}. From the same table, it is also evident that most of Europe is underrepresented. To address this lack of diversity, we carefully curate data across Europe, ranging from the snowy parts of northern Sweden to the sunny countryside of Italy. In total, we provide data across 14 different countries. To quantitatively evaluate the geographical diversity of our dataset, we use the diversity area metric defined in \cite{waymo_2020} as the union of all \si{75\meter} (radius) diluted ego-poses in the dataset. Using this definition, ZOD \textit{Frames} obtains an area metric of \si{705\kilo\meter\squared}, which could be compared to
\si{5\kilo\meter\squared}, \si{17\kilo\meter\squared}, and \si{76\kilo\meter\squared} for nuScenes~\cite{nuscenes_cvpr_2020},  Argoverse 2~\cite{argoverse2_2021}, and Waymo Open \cite{waymo_2020}, respectively\footnote{The geographical coverage values for the nuScenes and Waymo Open datasets are taken from \cite{waymo_2020}.}. Furthermore, we argue that  -- as our sensors and annotations range way beyond \si{75\meter} -- we can alter the definition to include the union of all \si{150\meter} radius circles around the ego-poses, in which case ZOD \textit{Frames} spans an area of \si{2039\kilo\meter\squared} which is $26\times$ the area covered in \cite{waymo_2020}. Importantly, we only include annotated frames when computing these metrics. We also show the entire geographical distribution of ZOD \textit{Frames} in \cref{fig:geographical-dist}. 

To further illustrate the diversity of ZOD, we analyze the distribution of \textit{Frames} across different weather conditions, road types, and times of day. As shown in \cref{fig:metadata_weather_roadtype_daytime}, our dataset includes frames captured in various weather conditions, including clear, cloudy, rainy, foggy, and snowy conditions. Moreover, our dataset covers different road types, including highways, urban roads, rural roads, and city streets, enabling the development and evaluation of models that can operate in different driving scenarios. Finally, we note that our dataset includes frames captured during different times of day, including almost 20k nighttime scenes, providing a comprehensive representation of real-world driving scenarios. By considering these diverse attributes during curation, we ensure that our dataset covers a broad range of driving conditions, enabling the development and evaluation of robust AD systems.

\subsection{Long-range perception}
\label{sec:long-range-objects}
As explained in \cref{sec:diversity}, ZOD is a highly diverse dataset containing data from various driving conditions, ranging from slow-moving city driving to high-speed highway driving. To operate a vehicle safely when driving at speeds up to \si{130\kilo\meter\per\hour} (maximum ego-vehicle speed in ZOD is \si{133\kilo\meter\per\hour}), it is crucial to detect objects not only in your vicinity, but also at longer distances. To accurately perceive the environment at distances required for high-speed driving, the ego-vehicle has to be equipped with sensors with sufficient resolution to enable long-range perception. In ZOD, we have an 8MP front-looking camera coupled with high-resolution LiDAR sensors, allowing annotation of objects up to 245 meters away. This is -- to the best of the authors' knowledge -- farther than any other publicly available AD dataset of comparable size. In \cref{fig:object-range-distribution}, we show the distribution of the distance to the annotated 3D objects for three top-level classes, namely vehicles, vulnerable vehicles, and pedestrians. Note that the classes of the other datasets have been mapped to these three top-level classes. In particular, ZOD exhibits a range distribution similar to the Argoverse 2 dataset, with the exception of having a longer tail for vehicles and a lower density for distant vulnerable vehicles.

\begin{table}[t]
    \centering
    \caption{3D object detection performance by range.}
    \resizebox{\linewidth}{!}{
    \begin{tabular}{c|c|c|c|c|c}
        & 0-150m & 0-50m & 50-100m & 100-150m & 150-250m \\ \hline
        mAP & 0.25 & 0.33 & 0.19 & 0.06 & 0.01 \\
        CDS & 0.46 & 0.62 & 0.33 & 0.08 & 0.02
    \end{tabular}
    }
    \label{tab:centerpoint-ranges}
    \vspace{-2mm}
\end{table}

\begin{table}[t]
\centering
    \caption{Traffic sign classification metrics [\%].}
    \begin{tabular}{c|c|c|c}
    $\text{F}_1^{\text{macro}}$ & $\text{F}_1^{\text{micro}}$ & $\text{Acc}_{\text{avg}}^{\uparrow10}$ & $\text{Acc}_{\text{avg}}^{\downarrow10}$  \\ 
    \hline
    78.5 & 95.4 & 93.4 & 65.4
    \end{tabular}
    \label{tab:tsr-classification}
\end{table}

We also aim to characterize the difficulty of 3D object detection on \textit{Frames} by employing the widely used CenterPoint \cite{yin2021center}. We trained CenterPoint and evaluated its performance across various range bins using the mAP and CDS metrics, as outlined in \cite{argoverse2_2021}\footnote{The 0-150m bin corresponds to Argoverse 2's evaluation range, where CenterPoint achieves mAP=0.18, compared to mAP=0.25 on ZOD}. A comprehensive breakdown of the results is presented in \cref{tab:centerpoint-ranges}. Our findings indicate that the detector struggles significantly at extended ranges, emphasizing the need for improved detectors and/or more relevant metrics. These challenges can be addressed using ZOD.

\subsection{Long-tail perception}
\label{sec:long-tail-perception}
\begin{figure}[t]
  \centering
   \includegraphics[width=\linewidth]{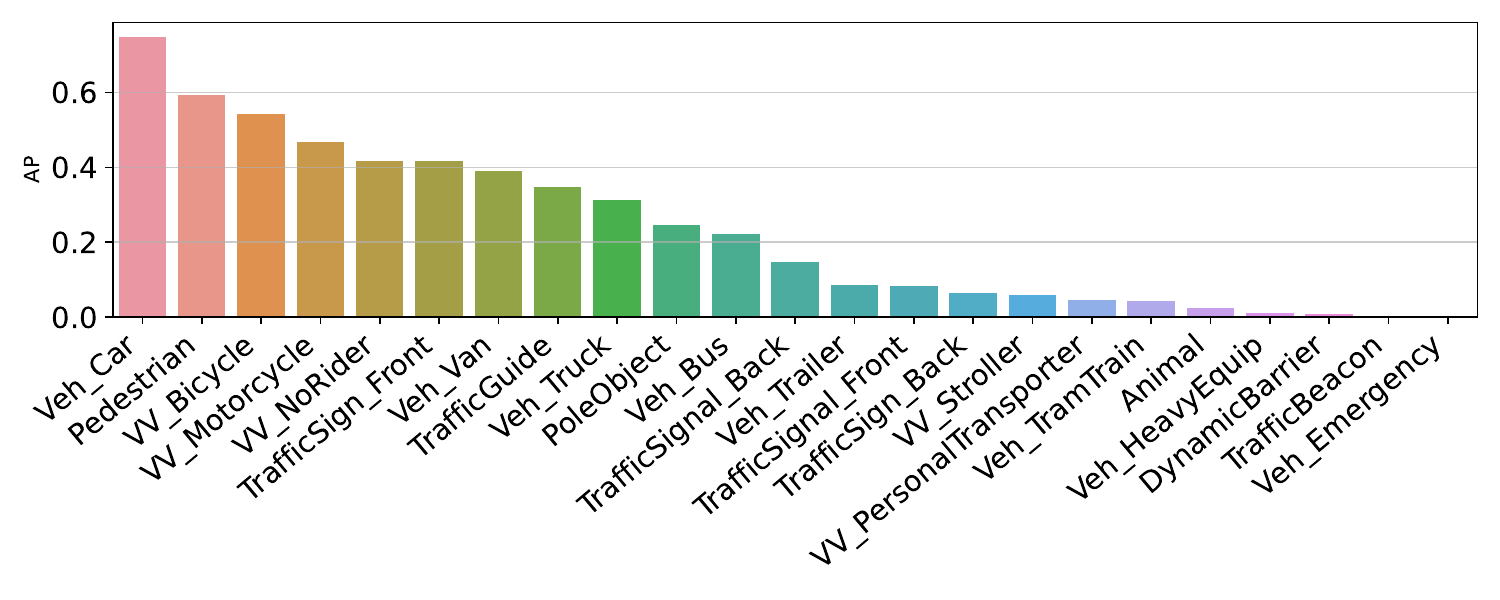}
   \vspace{-5mm}
   \caption{3D object detection AP per class (0-250m).}
   \label{fig:centerpoint_performance}
   \vspace{-2mm}
\end{figure}

A critical dimension of autonomous driving is managing infrequent and challenging situations, including the precise detection of uncommon objects like wheelchairs and strollers. In examining this, we evaluate CenterPoint's performance across an extensive array of classes over the full 250m distance, as illustrated in Figure \ref{fig:centerpoint_performance}. While cars are consistently detected, nearly half of the analyzed classes register an AP score of under 10\%. 

In the realm of traffic sign recognition (TSR), a significant challenge emerges from the marked class imbalance between commonly recognized, universal signs (\eg, pedestrian crossing) and those specific to rare scenarios (\eg, warning for polar bears). We implement a simple ResNet-50-based classification baseline for TSR on ZOD and report the results in \cref{tab:tsr-classification}. We exclude classes with less than 10 signs in the validation set. Common signs are classified very well, as shown by high F1-micro score, whereas the class-balanced F1-macro score is significantly lower. We also show accuracy for the 10 most common and 10 least common classes, which further highlight the difference in performance on common vs. rare classes. Jointly, these results warrants further investigation into long-tailed tasks to enhance the reliability of autonomous systems.


\vspace{-1mm}
\subsection{Anonymization}
\label{sec:anonymization_study}
To analyze the effects that anonymization has on downstream computer vision tasks, we train two image-based object detectors on three different versions of the images: the original, blurred, and DNAT images. Training is done on each of the anonymized image sets separately, while the evaluation is performed on the original images. We train on the three classes for which anonymization is relevant, namely vehicles, vulnerable vehicles (bicycles, motorcycles, wheelchairs, \etc), and pedestrians. The first detection pipeline is a Faster-RCNN \cite{ren2015faster}, coupled with a feature pyramid network  \cite{lin2017feature} and a Resnet-50 \cite{he2016deep} backbone -- implemented in the Detectron2 \cite{wu2019detectron2} framework -- while the second pipeline is YOLOv7 \cite{wang2022yolov7}. The results are computed according to the COCO evaluation protocol \cite{coco_eccv14} and presented in \cref{tab:2d-anonymization} (more comprehensive results are available in the supplementary material). The results show no statistically significant performance degradation when training with anonymized images over the original setting. Moreover, these results act as a baseline for image-based object detection on ZOD \textit{Frames}.

%% file: include_iccv_camready/conclusion.tex
\section{Conclusion}
\label{sec:conclusion_futurework}
We present ZOD, a diverse multimodal dataset for autonomous driving. ZOD contains data collected, with high-resolution sensors, from 14 European countries, thereby addressing the lack of European data in publicly available AD datasets. With this geographical diversity we are able to provide a wide range of driving scenarios, covering snowy country roads in northern Sweden, rainy highways in Germany, busy downtown traffic in France, and sunny suburban roads in Italy. We supply a comprehensive set of dense annotations, including semantic/instance segmentation masks for lane markings, road paintings, and ego road, 2D/3D bounding boxes for objects, and classification labels of the road surface condition. Notably, the 3D object annotations range up to 245 meters, which is farther than any comparable AD dataset. Additionally, ZOD boasts a comprehensive traffic sign taxonomy, with a greater number of unique instances than any other comparable datasets. In terms of future work, we will add temporal-consistent annotations for the entire duration of the sequences in ZOD \textit{Sequences}, two seconds of supporting camera frames for all ZOD \textit{Frames}, and more \textit{Sequences} and \textit{Drives} with supporting high definition maps. We hope our diverse dataset can inspire research that drives the field even further toward robust and safe AD. 

%% file: include_iccv_camready/new_supplementary.tex
\section{Supplementary material}
\label{sec:supplementary}

\subsection{Drives}
\begin{figure}[ht]
\centering
    \includegraphics[width=0.9\linewidth]{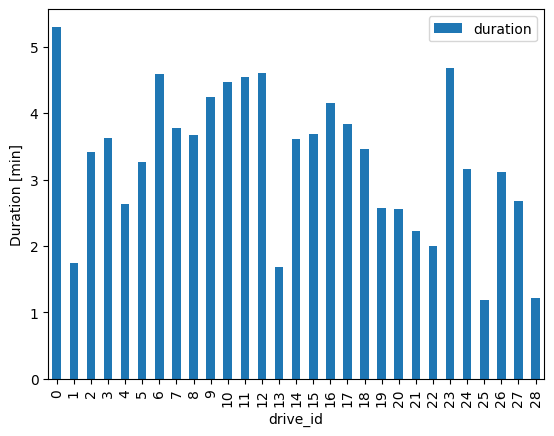}
    \caption{Duration of ZOD \textit{Drives}.}
    \label{fig:supp-drives-stats}
\end{figure}

\begin{figure}[ht]
\centering
\begin{subfigure}{0.48\linewidth}
    \centering
    \includegraphics[width=\linewidth]{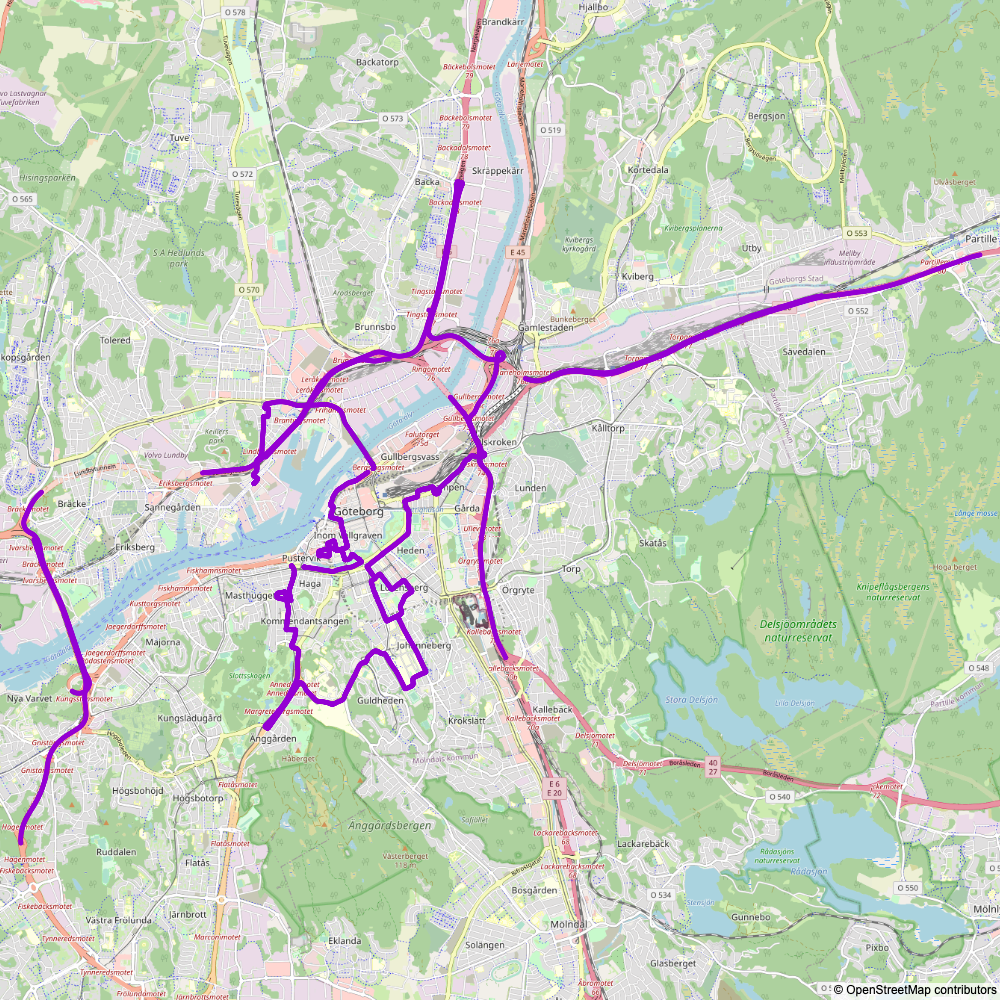}
    \caption{Gothenburg}
    \label{fig:drives_gps_gbg}
\end{subfigure}
\begin{subfigure}{0.48\linewidth}
    \centering
    \includegraphics[width=\linewidth]{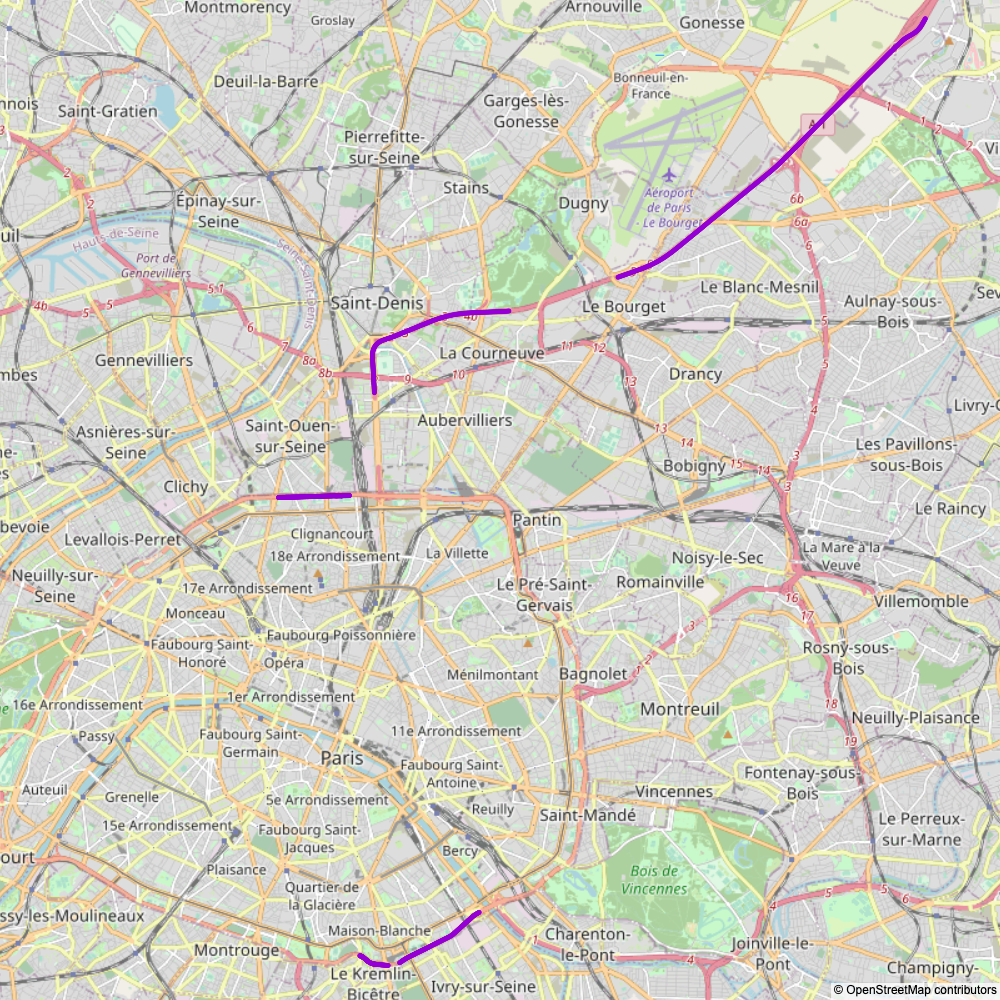}
    \caption{Paris}
    \label{fig:drives_gps_paris}
\end{subfigure}
    \caption{GNSS of ZOD \textit{Drives} projected onto the map.}
    \label{fig:supp-drives_gps}
\end{figure}

The 29 drives are collected in two cities, namely Gothenburg, Sweden and Paris, France. The average duration is 3 minutes, with the shortest being around 1 minute and the longest about 5 minutes. The combined duration is longer than 1 hour and 30 minutes. Further details can be seen in \cref{fig:supp-drives-stats} and \cref{fig:supp-drives_gps}.

\subsection{Annotation details}
\label{sup:annotation_details}
\begin{figure}[ht]
\centering
    \includegraphics[width=0.7\linewidth]{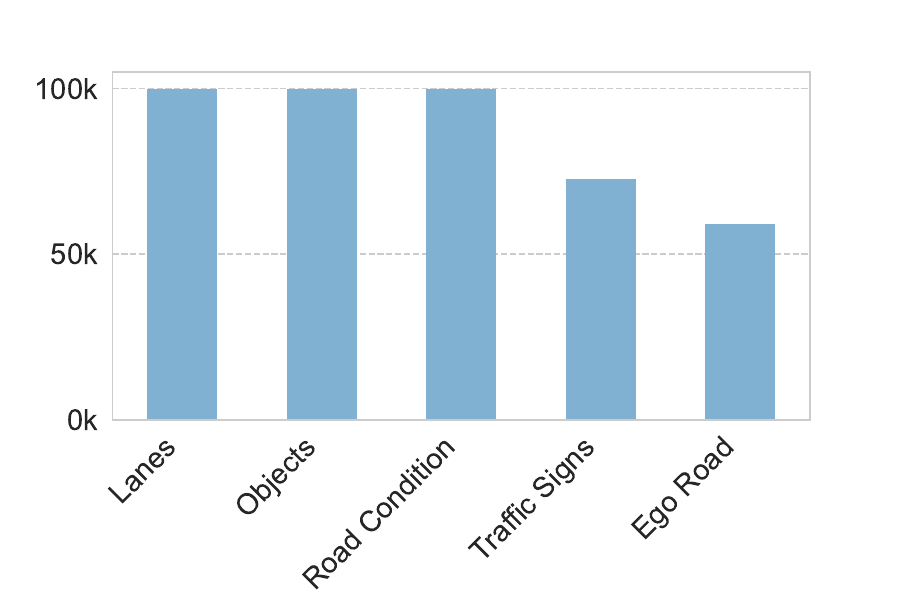}
    \caption{Number of annotated frames per project.}
    \label{fig:supp-project_counts}
\end{figure}

We provide complete multi-task annotations for \textit{Sequences}, but \textit{Frames} are only partially annotated for traffic sign and ego road annotations, as shown in ~\cref{fig:supp-project_counts}. All annotations are provided in the \textit{GeoJSON} format \cite{GeoJSON}, which encodes a variety of geometries in a \textit{JSON} format with additional properties. Annotation files can be easily read and visualized by the provided development kit. \cref{tab:annotation_details} details high-level classes, sub-classes, and additional attributes assigned to each annotated object or polygon for different annotation tasks in ZOD.

\subsection{Additional dataset statistics}
\begin{figure}[ht]
\vspace{-5mm}
\centering
    \includegraphics[width=0.85\linewidth]{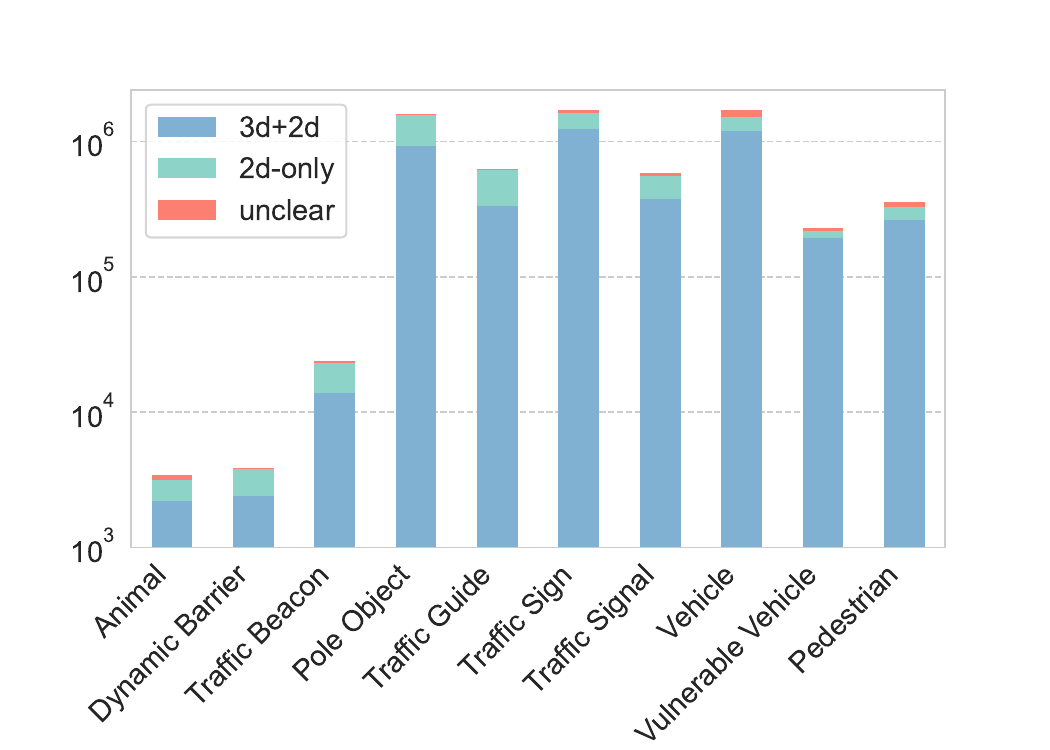}
    \caption{Number of annotated objects, broken down by class and annotation detail.}
    \label{fig:supp-object_counts}
\end{figure}

\begin{figure}[ht]
\centering
    \includegraphics[width=0.95\linewidth, trim={0, 10mm, 0, 0}, clip]{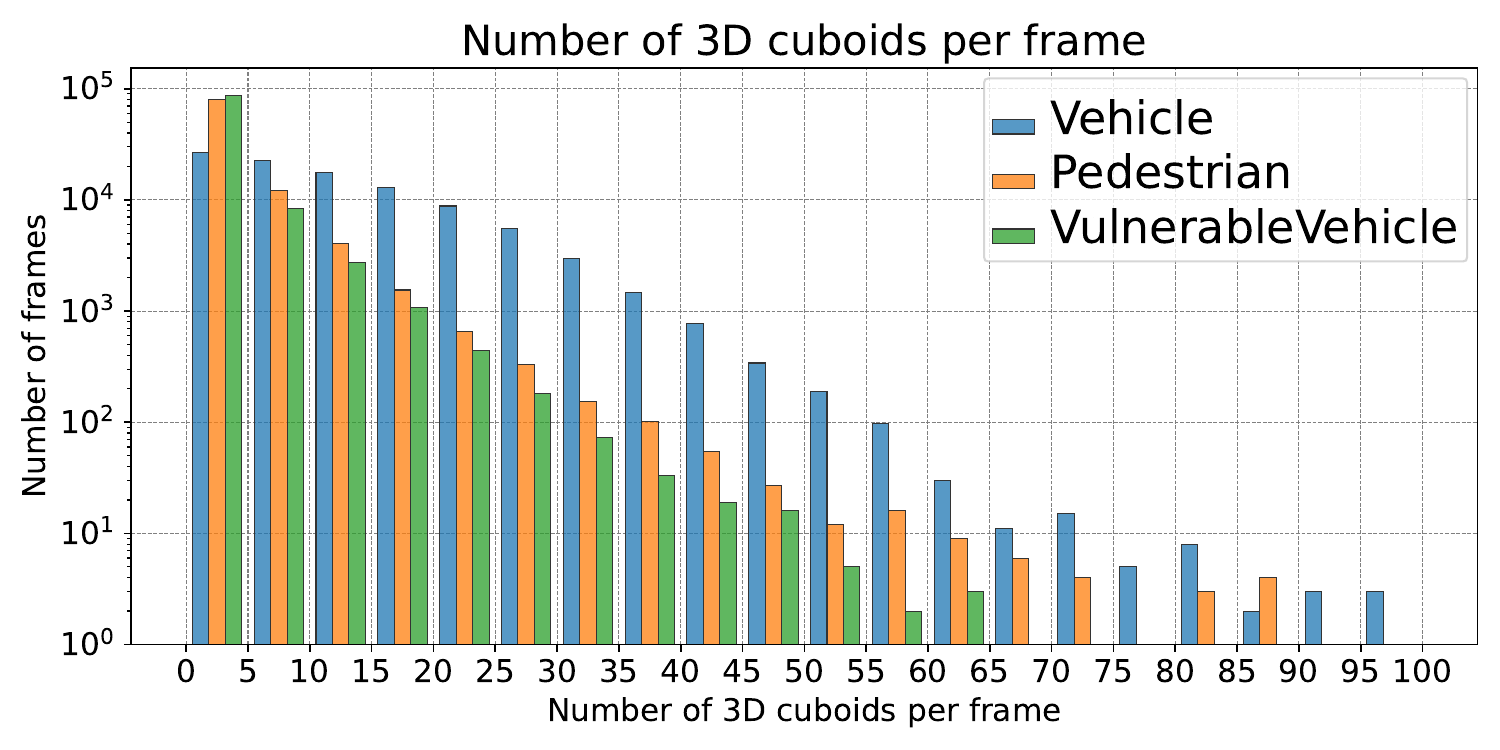}
    \vspace{-3mm}
    \caption{Annotated 3D cuboids per frame and class.}
    \label{fig:supp-3d_per_frame}
\end{figure}

We illustrate the number of annotated dynamic and static objects per class in ZOD \textit{Frames} in \cref{fig:supp-object_counts}. Note that some classes are much more represented than others. For example, vehicles are naturally orders of magnitude more prevalent than animals in traffic scenarios. The different colors represent the levels of detail that an object annotation can have, from paired 2D and 3D bounding boxes, to simply a 2D box, to a rough ``unclear'' region that indicates the possibility of one or more objects in the given image region.
We also show the number of annotated 3D cuboids per ZOD \textit{Frame} in \cref{fig:supp-3d_per_frame}, for three top-level classes -- namely -- Vehicle, Vulnerable Vehicle, and Pedestrian.

\begin{figure}[ht]
\vspace{-5mm}
\centering
    \includegraphics[width=0.75\linewidth]{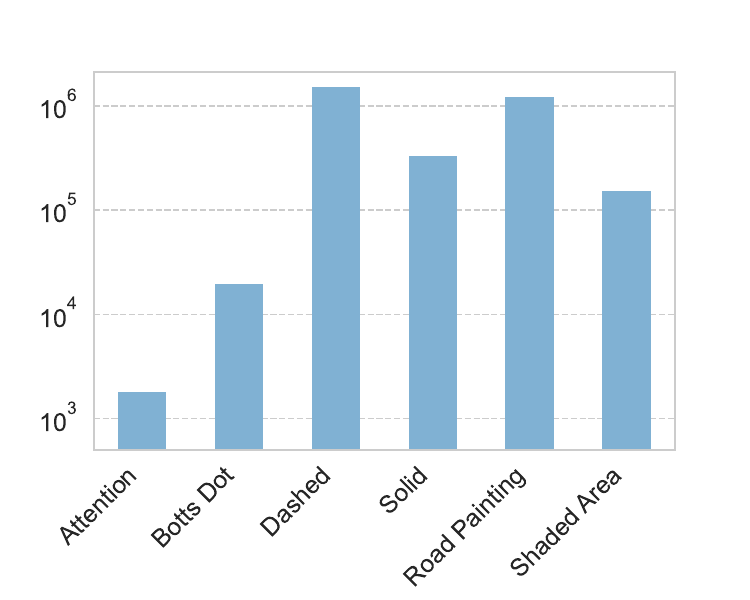}
    \vspace{-4mm}
    \caption{Number of annotated polygons for lanes, broken down by class, including road paintings and shaded areas.}
    \label{fig:supp-lane_counts}
\end{figure}

Similarly, we present the number of lane annotations per class in ZOD \textit{Frames} in~\cref{fig:supp-lane_counts}. These counts correspond to single polygons, not associated instances across many polygons. We note that ZOD contains many dashed and solid lines, as expected, but also a significant number of road paintings and shaded areas. The latter may be particularly of interest, since it corresponds to areas on the road where shadows are cast in a way that very strongly resembles lane markings. These are one of the primary causes for false positives, and having explicit annotations makes it possible to both train the network to be resistant to these distractors and also to perform target evaluations.

\begin{table*}[ht]
    \centering
    \resizebox{0.75\textwidth}{!}{%
        \begin{tabular}{ll|ccccccccc}
            \toprule
            Train                    & Eval & $AP$   & $AP50$ & $AP75$ & $AP_s$ & $AP_m$ & $AP_l$ & $AP_{veh}$ & $AP_{VV}$     & $AP_{ped}$ \\
            \midrule
            \multirow{3}{25pt}{blur}
                                         & blur     & 30.116 & 54.562 & 28.693 & 7.273  & 30.479 & 50.981 & 42.393     & 25.753        & 22.203     \\
                                         & DNAT     & 30.131 & 54.554 & 28.733 & 7.264  & 30.476 & 50.998 & 42.389     & 25.748        & 22.257     \\
                                         & original & 30.117 & 54.558 & 28.714 & 7.266  & 30.476 & 50.975 & 42.398     & 25.738        & 22.217     \\
            \midrule
            \multirow{3}{25pt}{DNAT}     & blur     & 30.277 & 54.988 & 28.870 & 7.210  & 30.645 & 51.126 & 42.452     & 26.028        & 22.350     \\
                                         & DNAT     & 30.305 & 54.927 & 28.912 & 7.200  & 30.662 & 51.255 & 42.458     & 26.050        & 22.406     \\
                                         & original & 30.308 & 54.931 & 28.918 & 7.208  & 30.657 & 51.281 & 42.463     & 26.046        & 22.416     \\
            \midrule
            \multirow{3}{25pt}{original} & blur     & 30.317 & 54.746 & 28.897 & 7.288  & 30.649 & 51.142 & 42.479     & 26.186        & 22.285     \\
                                         & DNAT     & 30.352 & 54.860 & 28.933 & 7.295  & 30.676 & 51.323 & 42.499     & 26.186 &	22.372              \\
                                         & original & 30.352 & 54.863 & 28.921 & 7.289  & 30.661 & 51.329 & 42.497     & 26.182        & 22.378     \\
            \bottomrule
        \end{tabular}
    }
    \caption{Experimental results when training and evaluating on all possible combinations of anonymization methods. }
    \label{tab:supp-anoymization}
\end{table*}

\subsection{Additional anonymization results}
To provide a more comprehensive view of the anonymization experiments, we provide the complete experimental results in \cref{tab:supp-anoymization}. Here, we show the performance of the Faster-RCNN \cite{ren2015faster} pipeline when trained and evaluated across all possible combinations of anonymization methods.

\subsection{Compression results}
\label{sec:png_jpg_comparison}
ZOD contains high-resolution images of every scene. For each image, we do JPG compression of the original PNG image to reduce the memory footprint of ZOD, increasing the usability of the dataset. As the JPG compression is lossy, we analyze the effect this has on downstream tasks. To evaluate this effect, we train the Faster-RCNN object detection pipeline on both JPG and PNG images, and see no significant performance degradation due to compression when evaluating both trained networks on the PNG images, see \cref{tab:supp-png-jpg}.

\begin{table}[h]
    \centering
    \vspace{-2mm}
    \resizebox{0.3\textwidth}{!}{%
    \begin{tabular}{lcc}
        \toprule
        Metric     & PNG              & JPG              \\
        \midrule
        $AP$       & 30.23 $\pm$ 0.09 & 30.02 $\pm$ 0.03 \\
        $AP50$     & 54.79 $\pm$ 0.06 & 54.34 $\pm$ 0.10 \\
        $AP75$     & 28.72 $\pm$ 0.15 & 28.51 $\pm$ 0.02 \\
        \midrule
        $AP_s$     & 7.23 $\pm$ 0.04  & 7.14 $\pm$ 0.06  \\
        $AP_m$     & 30.49 $\pm$ 0.14 & 30.30 $\pm$ 0.04 \\
        $AP_l$     & 51.23 $\pm$ 0.07 & 50.95 $\pm$ 0.02 \\
        \midrule
        $AP_{veh}$ & 42.41 $\pm$ 0.07 & 42.26 $\pm$ 0.04 \\
        $AP_{VV}$  & 25.96 $\pm$ 0.15 & 25.71 $\pm$ 0.06 \\
        $AP_{ped}$ & 22.32 $\pm$ 0.04 & 22.08 $\pm$ 0.07 \\
        \bottomrule
    \end{tabular}
    }
    \caption{Impact of image compression. We report AP (computed according to COCO evaluation protocol \cite{coco_eccv14}) when training Faster-RCNN on compressed vs. uncompressed images. The metrics are computed using the original (uncompressed) images, and are presented as the mean and standard deviation across three separate runs.}
    \label{tab:supp-png-jpg}
\end{table}

\subsection{Panoptic segmentation benchmark}
We use Panoptic-DeepLab \cite{cheng2020panoptic} as the panoptic segmentation baseline. Here, we set lanes as \textit{things} while ego-road and cross-walks are set as \textit{stuff}. \cref{tab:panoptic-results} shows Panoptic-Quality (PQ), segmentation quality component (SQ), and recognition quality component (RQ). For context, SOTA PQ results for the BDD100K \cite{bdd100k_2018} and CityScapes \cite{cityscape_cvpr16} datasets are 23.90 and 70.1, respectively.
\vspace{-1mm}
\begin{table}[ht]
    \centering
    \caption{Panoptic segmentation}
     \vspace{-2mm}
    \begin{tabular}{l|ccc}
        Class & PQ & SQ & RQ \\ \hline
        \textit{Things} & 40.0 & 77.6  & 51.5  \\
        \textit{Stuff} & 67.8 & 88.3  & 72.7 \\
        All & 60.9 & 85.6  & 67.6 
    \end{tabular}
    \label{tab:panoptic-results}
\end{table}

\begin{table*}[ht]
    \centering
    \resizebox{0.95\textwidth}{!}{%
        \begin{tabular}{l|c|c|c}
            \toprule
            Annotation tasks              			& High-level classes                       & Sub-classes                          		& Additional attributes     \\
            \midrule
            \multirow{2}{*}{Ego road}       	& Road                                      		 & N/A                                				  & N/A   \\
            \cmidrule{2-4}
                                                               & Debris                                   		   & N/A                                 				& N/A   \\
            \midrule
            \multirow{5}{*}{\begin{tabular}[c]{@{}l@{}}Lane markings\\ and\\ road paintings\end{tabular}}
            & Solid                                     & N/A                                				    & \multirow{3}{*}{InstanceID, Coloured, MultipleLaneMarkings}  \\
            \cmidrule{2-3}
            & Dashed                                 & N/A                                					&																									\\
            \cmidrule{2-3}
            & Botts dot                              & N/A                                					 &																									 \\
            \cmidrule{2-4}
            & Shaded area                         & Split, Merge, Undefined            		 & N/A   \\
            \cmidrule{2-4}
            & Road paintings                      & {\begin{tabular}[c]{@{}c@{}c@{}c@{}c}ContainsArrow, ContainsPictogram, ContainsMarker, \\ContainsTrafficSigns, ContainsCrossWalks, \\ContainsText, ContainsOther, Odd, Unclear\end{tabular}}                                                     & N/A   \\
            \midrule
            \multirow{10}{*}{\begin{tabular}[c]{@{}l@{}}Dynamic\\ and\\ static objects\end{tabular}}
            & Vehicle                                 & {\begin{tabular}[c]{@{}c}Car, Van, Truck, Trailer, Bus, \\HeavyEquip, TramTrain, Other, Inconclusive\end{tabular}}      			  & {\begin{tabular}[c]{@{}c}Unclear, OcclusionRatio, Emergency,\\RelativePosition, IsPullingOrPushing\end{tabular}}   \\
            \cmidrule{2-4}
            & Vulnerable vehicle               &  {\begin{tabular}[c]{@{}c}Bicycle, Motorcycle, Wheelchair, Stroller,\\PersonalTransporter, Other, Inconclusive\end{tabular}}      & {\begin{tabular}[c]{@{}c}Unclear, OcclusionRatio, Emergency,\\IsPullingOrPushing, WithRider\end{tabular}}   \\
            \cmidrule{2-4}
            & Pedestrian                           & N/A                                & {\begin{tabular}[c]{@{}c}Unclear, OcclusionRatio, Emergency, \\RelativePosition, IsPullingOrPushing\end{tabular}}   \\
            \cmidrule{2-4}
            & Animal                                  & N/A                                & N/A   \\
            \cmidrule{2-4}
            & TrafficSign                           & See below for more granular categorization of Traffic signs                                  & {\begin{tabular}[c]{@{}c@{}c@{}c@{}c@{}c}Unclear, OcclusionRatio, InOnDynamicObject, \\IsForConstruction, TrafficContentVisible
            \\IsForEgoRoad, IsForEgoLane, \\IsForOtherTrafficParticipants, \\ComplementaryToLandmark, IsSticker\end{tabular}}   \\
            \cmidrule{2-4}
            & TrafficSignal                        & N/A                                & {\begin{tabular}[c]{@{}c@{}c@{}c@{}c}Unclear, OcclusionRatio, InOnDynamicObject, \\IsForConstruction, TrafficContentVisible
            \\IsForEgoRoad, IsForEgoLane, \\IsForOtherTrafficParticipants\end{tabular}}   \\
            \cmidrule{2-4}
            & TrafficGuide                        & {\begin{tabular}[c]{@{}c@{}c}Reflector, Attention, SnowMarker, \\Bollard, Other, Inconclusive\end{tabular}}				 & {\begin{tabular}[c]{@{}c}Unclear, OcclusionRatio, InOnDynamicObject,\\ IsForConstruction, TrafficContentVisible, IsSticker\end{tabular}}   \\
            \cmidrule{2-4}
            & PoleObject                          & {\begin{tabular}[c]{@{}c@{}c}LampPole, LandmarkPole, \\LargeLandmarkPole, Other, Inconclusive\end{tabular}}		  & \multirow{3}{*}{{\begin{tabular}[c]{@{}c@{}c}Unclear, OcclusionRatio, InOnDynamicObject, \\IsForConstruction, TrafficContentVisible\end{tabular}}}   \\
            \cmidrule{2-3}
            & TrafficBeacon                      & N/A                                &    \\
            \cmidrule{2-3}
            & DynamicBarrier                    & N/A                                &    \\
            \midrule
            \multirow{8}{*}[0em]{Traffic signs}
            & MandatorySigns                   & {\begin{tabular}[c]{@{}c@{}c@{}c@{}c@{}c}PassOnThisSideLeft, PassOnThisSideRight, \\PassOnEitherSide, ProceedStraightOrTurnLeft, \\ProceedStraight, ProceedStraightOrTurnRight, \\TurnLeftAhead, TurnRightAhead, TurnAhead, \\TurnRight, TurnLeft, Roundabout\end{tabular}}                                & \multirow{7}{*}[-9em]{{\begin{tabular}[c]{@{}c@{}c@{}c@{}c@{}c@{}c}Unclear, OcclusionRatio, \\ComplementaryToLandmark, ContainsInnerSigns, \\ContentContainsText, IsForConstruction, \\IsForEgoRoad, IsForEgoLane, IsElectronic\\IsForOtherTrafficParticipants, \\IsOnDynamicObjects \end{tabular}}}  \\
            \cmidrule{2-3}
            & PrioritySigns                         & {\begin{tabular}[c]{@{}c@{}c@{}c}GiveWay,  GiveWayOncoming, \\PriorityStop, PrioOverOncoming, \\PriorityRoadBegin, PriorityRoadEnd\end{tabular}}       &   \\
            \cmidrule{2-3}
            & ProhibitorySigns                   & {\begin{tabular}[c]{@{}c@{}c@{}c@{}c@{}c}NoEntry, NoParking, NoStopping, \\NoUTurn, NoTurn, RoadClosed, \\NoOvertakingBegin, NoOvertakingEnd, \\MaximumSpeedLimitXBegin, MaximumSpeedLimitXEnd, \\SpeedLimitZoneXBegin, SpeedLimitZoneXEnd\end{tabular}}      &   \\
            \cmidrule{2-3}
            & RoadTypeSigns                    & {\begin{tabular}[c]{@{}c} MotorwayBegin, MotorwayEnd\end{tabular}}      &   \\
            \cmidrule{2-3}
            & SpecialSigns                         & {\begin{tabular}[c]{@{}c@{}c}VulnurableRoadUserCrossing, \\VulnurableRoadUserPathWay, \\IndicationCameraSurveillance \end{tabular}}      &   \\
            \cmidrule{2-3}
            & WarningSigns                       & {\begin{tabular}[c]{@{}c@{}c@{}c@{}c@{}c@{}c@{}c}Children, Crossing, Cyclists, Animal, \\Curve, RoadWorkBegin, RoadWorkEnd, \\ Roundabout, TrafficSignalAhead, \\ RoadNarrows, RoadBump, RoughRoad, \\ Slippery, GenericWarning, CongestionAhead, \\TwoWayTraffic, MergingTraffic, Crossroads, \\DoubleCurve, TunnelAhead \end{tabular}}     &   \\
            \cmidrule{2-3}
            & NotListed                             & N/A                               &    \\
            \cmidrule{2-4}
            & Unclear                                & N/A                              &  N/A \\
            \midrule
            \multirow{1}{*}{Road condition}
            & N/A                                      & N/A                            & Wetness, SnowCoverage   \\
            \bottomrule
        \end{tabular}
    }
    \caption{Annotation details for different tasks in ZOD.}
    \label{tab:annotation_details}
\end{table*}